# A Variational Approach for Approximating Bayesian Networks by Edge Deletion

**Arthur Choi** and **Adnan Darwiche**
Computer Science Department
University of California, Los Angeles
Los Angeles, CA 90095
{*aychoi,darwiche*}*@cs.ucla.edu*

## Abstract

We consider in this paper the formulation of approximate inference in Bayesian networks as a problem of exact inference on an approximate network that results from deleting edges (to reduce treewidth). We have shown in earlier work that deleting edges calls for introducing auxiliary network parameters to compensate for lost dependencies, and proposed intuitive conditions for determining these parameters. We have also shown that our earlier method corresponds to Iterative Belief Propagation (IBP) when enough edges are deleted to yield a polytree, and corresponds to some generalizations of IBP when fewer edges are deleted. In this paper, we propose a different criteria for determining auxiliary parameters based on optimizing the KL–divergence between the original and approximate networks. We discuss the relationship between the two methods for selecting parameters, shedding new light on IBP and its generalizations. We also discuss the application of our new method to approximating inference problems which are exponential in constrained treewidth, including MAP and nonmyopic value of information.

## 1 INTRODUCTION

The complexity of algorithms for exact inference on Bayesian networks is generally exponential in the network treewidth (Jensen, Lauritzen, & Olesen, 1990; Lauritzen & Spiegelhalter, 1988; Zhang & Poole, 1996; Dechter, 1996; Darwiche, 2001). Therefore, networks with high treewidth (and no local structure, Chavira & Darwiche, 2005) can be inaccessible to these methods, necessitating the use of approximate algorithms. Iterative Belief Propagation (IBP), also known as Loopy Belief Propagation (Pearl, 1988; Murphy, Weiss, & Jordan, 1999), is one such algorithm that has been critical for enabling certain classes of applications, which have been intractable for exact algorithms (e.g., Frey & MacKay, 1997). We have proposed in previous work a new perspective on this influential algorithm, viewing it as an exact inference algorithm on a polytree approximation of the original network (Choi & Darwiche, 2006; Choi, Chan, & Darwiche, 2005). The approximate polytree results from deleting edges from the original network, where the loss of each edge is offset by introducing new parameters into the approximate network. We have shown that the iterations of IBP can be understood as searching for specific values of these parameters that satisfy intuitive conditions that we characterized formally (Choi & Darwiche, 2006). This has led to a number of implications. On the theoretical side, it provided a new, network–specific, characterization of the fixed points of IBP. On the practical side, it has led to a concrete framework for improving approximations returned by IBP by deleting fewer edges than those necessary to yield a polytree; that is, we delete enough edges to obtain a multiply connected network which is still tractable for exact inference.

In this paper, we consider another criterion for determining the auxiliary parameters introduced by deleting edges, which is based on minimizing the KL–divergence between the original and approximate network. This proposal leads to a number of interesting results. First, we provide intuitive, yet necessary and sufficient, conditions that characterize the stationary points of this optimization problem. These conditions suggest an iterative procedure for finding parameters that satisfy these conditions, leading to a new approximate inference method that parallels IBP and its generalizations. Second, the sufficiency of these conditions lead to new results on IBP and its generalizations, characterizing situations under which these algorithms will indeed be optimizing the KL–divergence.

We seek to optimize the form of the KL–divergence that uses weights from the original distribution, and as it turns out, the update equations for our new method are more expensive than those for IBP and its generalizations, requiring the availability of true node marginals in the original network. This means that the method as described is, in general, applicable only to networks whose treewidth is manageable, but whose constrained treewidth is not.[1] That is, this approximation will typically be useful for problems which remain hard even if treewidth is manageable. This includes MAP (Park & Darwiche, 2004), inference in credal networks (Cozman, de Campos, Ide, & da Rocha, 2004), and the computation of nonmyopic value of information (Krause & Guestrin, 2005). In complexity theoretic terms, computing node marginals is PP–complete, while computing MAP is $\mathrm{NP}^{\mathrm{PP}}$–complete. Hence, our proposed method can be used to approximate $\mathrm{NP}^{\mathrm{PP}}$–complete problems, while IBP and its generalizations approximate PP–complete problems.

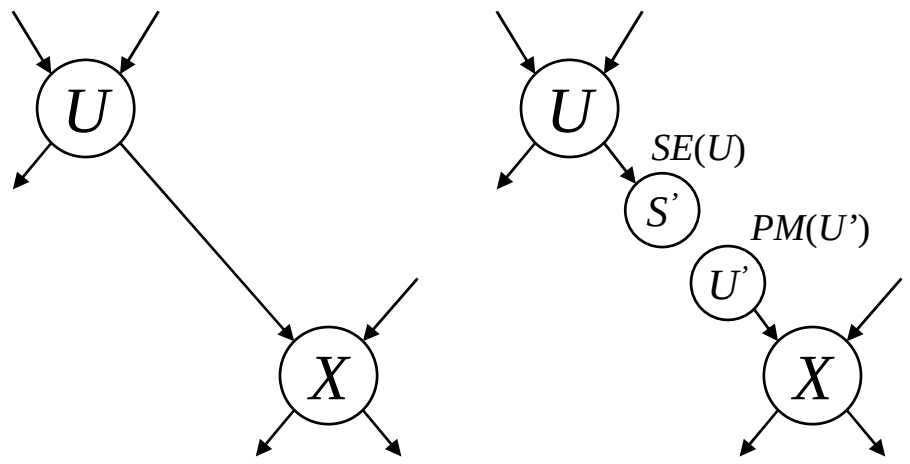

Figure 1: Deleting edge $U \to X$ by adding a clone $U'$ of $U$ and a binary evidence variable $S'$.

This paper is structured as follows. Section 2 reviews the framework of approximating networks by edge deletion. Section 3 treats the characterization of auxiliary parameters introduced by deleting edges, discussing the new characterization proposed in this paper, and comparing it to the one corresponding to IBP and its generalizations. Section 4 considers the problem of selecting which edges to delete in order to optimize the quality of approximations. Section 5 presents empirical results, Section 6 discusses related work, and Section 7 closes with some concluding remarks. Proofs of theorems are sketched in Appendix A.

## 2 DELETING AN EDGE

Let $U \to X$ be an edge in a Bayesian network, and suppose that we wish to delete this edge to make the network more amenable to exact inference algorithms. This deletion will introduce two problems. First, variable $X$ will lose its direct dependence on parent $U$. Second, variable $U$ may lose evidential information received through its child $X$. To address these problems, we propose to add two auxiliary variables for each deleted edge $U \to X$ as given in Figure 1. The first is a variable $U'$ which is made a parent of $X$, therefore acting as a clone of the lost parent $U$. The second is an *instantiated* variable $S'$ which is made a child of $U$, meant to provide evidence on $U$ in lieu of the lost evidence.[2] Note that the states $u'$ of auxiliary variable $U'$ are the same as the states $u$ of variable $U$, since $U'$ is a clone of $U$. Moreover, auxiliary variable $S'$ is binary as it represents evidence.

The deletion of an edge $U \to X$ will then lead to introducing new parameters into the network, as we must now provide conditional probability tables (CPTs) for the new variables $U'$ and $S'$. Variable $U'$, a root node in the network, needs parameters $\theta_{u'}$ representing the prior marginal on variable $U'$. We will use $PM(U')$ to denote these parameters, where $PM(u') = \theta_{u'}$. Variable $S'$, a leaf node in the network, needs parameters $\theta_{s'|u}$ representing the conditional probability of $s'$ given $U$. We will use $SE(U)$ to denote these parameters, where $SE(u) = \theta_{s'|u}$. Moreover, we will collectively refer to $PM(U')$ and $SE(U)$ as *edge parameters.* Figure 2 depicts a simple network with a deleted edge, together with one possible assignment of the corresponding edge parameters.

We have a number of observations about our proposal for deleting edges:

- The extent to which this proposal is successful will depend on the specific values used for the parameters introduced by deleting edges. This is a topic which we address in the following section.
- If the deleted edge $U \to X$ splits the network into two disconnected networks, one can always find edge parameters which are guaranteed to lead to exact computation of variable marginals in both subnetworks (Choi & Darwiche, 2006).
- The auxiliary variable $S'$ can be viewed as injecting a *soft evidence* on variable $U$, whose strength is defined by the parameters $SE(U)$. Note that for queries that are conditioned on evidence $s'$, only the relative ratios of parameters $SE(U)$ matter, not their absolute values (Pearl, 1988; Chan & Darwiche, 2005).

Our goal now is to answer the following two questions. First, how do we parametrize deleted edges? Second, which edges do we delete?

[1] Networks may admit elimination orders with manageable treewidths, but certain queries may constrain these orders, leading to constrained treewidths.

[2] Our proposal for deleting an edge is an extension of the proposal given by (Choi et al., 2005), who proposed the addition of a clone variable $U'$ but missed the addition of evidence variable $S'$.

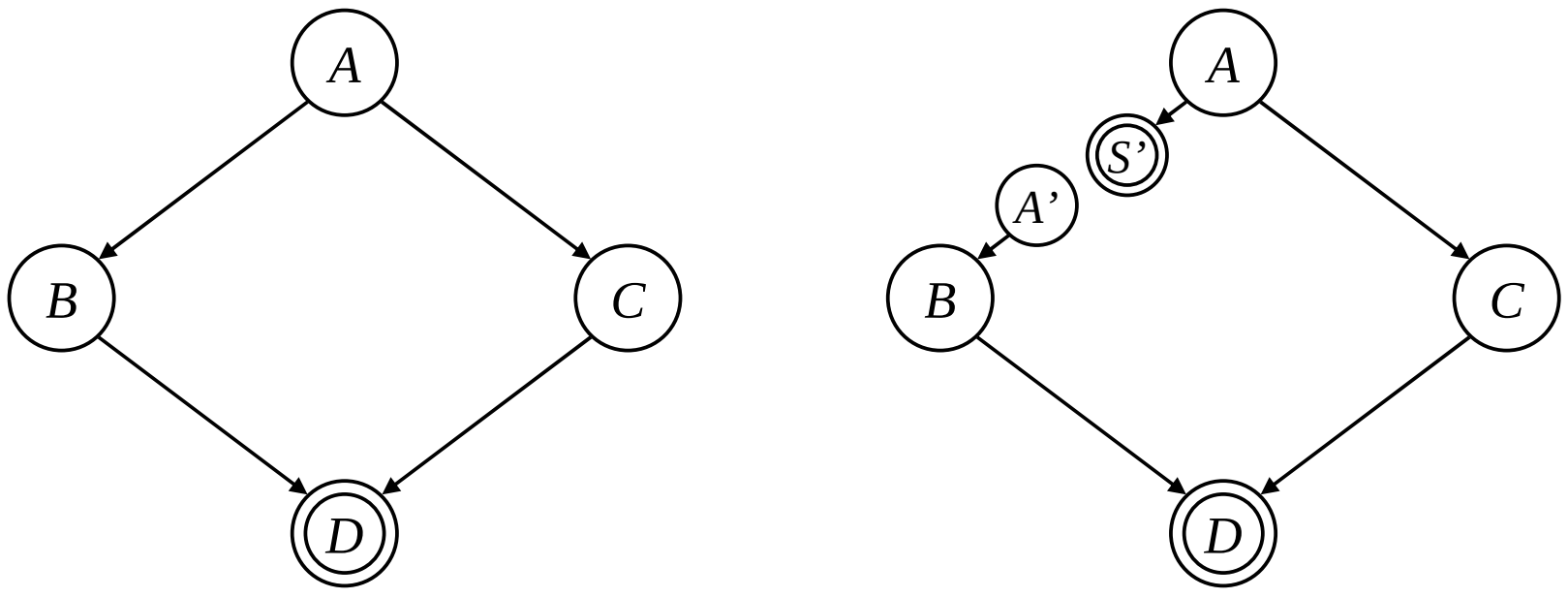

| $\Theta_{s'\|A}$ | $SE(A)$ |
|---|---|
| $\theta_{s'\|a}$ | 0.3437 |
| $\theta_{s'\|\bar{a}}$ | 0.6562 |

| $\Theta_{A'}$ | $PM(A')$ |
|---|---|
| $\theta_{a'}$ | 0.8262 |
| $\theta_{\bar{a}'}$ | 0.1737 |

Figure 2: A network $N$ (left), an approximate network $N'$ found after deleting $A \rightarrow B$ (center), along with parameters for auxiliary evidence variable $S'$ and clone $A'$ (right).

# 3 PARAMETRIZING EDGES

Given a network $N$ and evidence $\mathbf{e}$, our proposal is then to approximate this network with another $N'$ that results from deleting some edges $U \rightarrow X$ as given earlier. Moreover, when performing inference on network $N'$, we will condition on the augmented evidence $\mathbf{e}'$, composed of the original evidence $\mathbf{e}$ and all auxiliary evidence $s'$ introduced when deleting edges. More formally, if $Pr$ and $Pr'$ are the distributions induced by networks $N$ and $N'$, respectively, we will use $Pr'(\mathbf{X}|\mathbf{e}')$ to approximate $Pr(\mathbf{X}|\mathbf{e})$ where $\mathbf{X}$ is a set of variables in the original network $N$.

To completely specify our approximate network $N'$, we need to specify parameters $PM(u')$ and $SE(u)$ for each edge that we delete. We have proposed in (Choi & Darwiche, 2006) an iterative procedure that uses the following update equations to parametrize edges:

$$\begin{aligned} PM(u') &= \alpha \frac{\partial Pr'(\mathbf{e}')}{\partial \theta_{s'|u}} \\ SE(u) &= \alpha \frac{\partial Pr'(\mathbf{e}')}{\partial \theta_{u'}}, \end{aligned} \quad (1)$$

where $\alpha$ is a normalizing constant.[3] This procedure, which we call ED-BP, starts with some arbitrary values for $PM(U')$ and $SE(U)$, leading to an initial approximate network $N'$. This network can then be used to compute new values for these parameters according to the update equations in (1). The process is then repeated until convergence to a fixed point (if at all). We have also shown that when deleting enough edges to yield a polytree, the parametrizations $PM(U')$ and $SE(U)$ computed in each iteration correspond precisely to the messages passed by IBP. Moreover, if the edges deleted do not yield a polytree, ED-BP corresponds to a generalization of IBP (simulated by a particular choice of a joingraph; see also Aji & McEliece, 2001; Dechter, Kask, & Mateescu, 2002).

[3]This is an alternative, but equivalent, formulation of the update equations given by (Choi & Darwiche, 2006).

Finally, these update equations lead to fixed points characterized by the following conditions:

$$\begin{aligned} Pr'(u|\mathbf{e}') &= Pr'(u'|\mathbf{e}'), \\ Pr'(u|\mathbf{e}' \setminus s') &= Pr'(u'). \end{aligned} \quad (2)$$

The first condition says that variables $U'$ and $U$ should have the same posterior marginals. The second condition, in light of the first, says that the impact of evidence $s'$ on variable $U$ is equivalent to the impact of all evidence on its clone $U'$. Indeed, these conditions correspond to the intuitions that motivated ED-BP.

## 3.1 A VARIATIONAL APPROACH

We propose now a variational approach to parametrizing deleted edges, based on the KL–divergence:

$$KL(Pr(.|\mathbf{e}), Pr'(.|\mathbf{e}')) \stackrel{def}{=} \sum_w Pr(w|\mathbf{e}) \log \frac{Pr(w|\mathbf{e})}{Pr'(w|\mathbf{e}')},$$

where $w$ is a world, denoting an instantiation over all variables. Note that the KL–divergence is not symmetric: the divergence $KL(Pr(.|\mathbf{e}), Pr'(.|\mathbf{e}'))$ is weighted by the true distribution while the divergence $KL(Pr'(.|\mathbf{e}'), Pr(.|\mathbf{e}))$ is weighted by the approximate one. Common practice weighs the KL–divergence using the approximate distribution, which is typically more accessible computationally (e.g., Yedidia, Freeman, & Weiss, 2005). In contrast, we will weigh by the true distribution in what follows.

Before we proceed to optimize the KL–divergence, we must ensure that the domains of the distributions $Pr(.|\mathbf{e})$ and $Pr'(.|\mathbf{e}')$ coincide. One way to ensure this is to use the following construction, demonstrated in Figure 3. Given a Bayesian network $N^\star$, we can replace each edge $U \rightarrow X$ to be deleted with a chain $U \rightarrow U' \rightarrow X$, where the equivalence edge $U \rightarrow U'$ denotes an equivalence constraint: $\theta_{u'|u} = 1$ iff $u' = u$. The resulting augmented network $N$ will then satisfy three important properties. First, it is equivalent to the original network $N^\star$ over common variables. Second, it has the same treewidth as $N^\star$. Finally, when we delete equivalence edges $U \rightarrow U'$ from $N$, we get an approximate network $N'$ that does not require the introduction of clone variables $U'$ as they are already present in $N'$.[4] We can therefore compute the KL–divergence between the augmented network $N$ and its approximation $N'$.

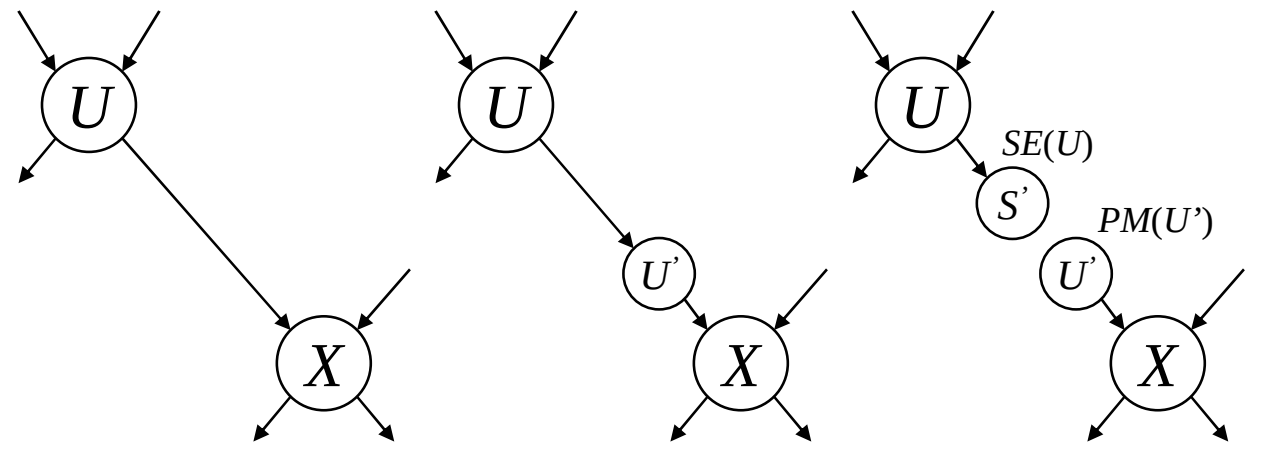

Figure 3: The edge $U \rightarrow X$ in $N^\star$ is replaced with a chain $U \rightarrow U' \rightarrow X$ in $N$. We delete the equivalence edge $U \rightarrow U'$ in $N$ to get $N'$.

We can now state the following key result:

**Lemma 1** *Let $N$ be an (augmented) Bayesian network and $N'$ be the network that results from deleting equivalence edges $U \rightarrow U'$. Then:*

$$KL(Pr(.|\mathbf{e}), Pr'(.|\mathbf{e}')) = \sum_{U \rightarrow U'} \sum_{u=u'} Pr(uu'|\mathbf{e}) \log \frac{1}{\theta_{u'}\theta_{s'|u}} + \log \frac{Pr'(\mathbf{e}')}{Pr(\mathbf{e})}.$$

Since $Pr'(\mathbf{e}')$ is a function of our edge parameters, the KL–divergence is thus also a function of our edge parameters. This result will be used later to derive update equations for our variational method, and to develop a heuristic for choosing edges to delete.

Before we proceed though, we observe the following. Let $\mathbf{X}$ be the variables of the original network $N^\star$ and $\mathbf{U}'$ be the clone variables introduced via the equivalence edges $U \rightarrow U'$ in $N$. The KL–divergence of Lemma 1 is then over the variables $\mathbf{XU}'$. One would normally prefer to optimize the KL–divergence $KL(Pr(\mathbf{X}|\mathbf{e}), Pr'(\mathbf{X}|\mathbf{e}'))$ over the original variables $\mathbf{X}$, but our method will seek to optimize $KL(Pr(\mathbf{XU}'|\mathbf{e}), Pr'(\mathbf{XU}'|\mathbf{e}'))$ instead. In fact, the properties of the KL–divergence tell us that

$$KL(Pr(\mathbf{X}|\mathbf{e}), Pr'(\mathbf{X}|\mathbf{e}')) \leq KL(Pr(\mathbf{XU}'|\mathbf{e}), Pr'(\mathbf{XU}'|\mathbf{e}')). \quad (3)$$

In our experimental results in Section 5, we will report results on both versions of the KL–divergence, referring to $KL(Pr(\mathbf{X}|\mathbf{e}), Pr'(\mathbf{X}|\mathbf{e}'))$ as the *exact KL,* and to $KL(Pr(\mathbf{XU}'|\mathbf{e}), Pr'(\mathbf{XU}'|\mathbf{e}'))$ as the *KL bound.*

[4] We still need to add a new child $S'$ for each $U$ however. Since variables $S'$ are observed, they do not prohibit us from computing the KL–divergence between $N$ and $N'$ even though they are not present in network $N$.

For the remainder of this paper, we will only be dealing with augmented networks $N$, leaving the original network $N^\star$ implicit.

## 3.2 THE APPROXIMATE NETWORK

We have the KL–divergence $KL(Pr(.|\mathbf{e}), Pr'(.|\mathbf{e}'))$ as a function of our edge parameters $PM(u') = \theta_{u'}$ and $SE(u) = \theta_{s'|u}$. If we set to zero the partial derivatives of the KL–divergence with respect to each edge parameter, we get the following.

**Theorem 1** *Let $N$ be a Bayesian network and $N'$ be the network that results from deleting equivalence edges $U \rightarrow U'$. The edge parameters of $N'$ are a stationary point of $KL(Pr(.|\mathbf{e}), Pr'(.|\mathbf{e}'))$ if and only if*

$$Pr'(u|\mathbf{e}') = Pr'(u'|\mathbf{e}') = Pr(u|\mathbf{e}), \quad (4)$$

*for all deleted edges $U \rightarrow U'$.*

That is, if we delete the edge $U \rightarrow U'$, then the marginals on both $U$ and $U'$ must be exact in the approximate network $N'$. Note, however, that this does not imply that other node marginals must be exact in $N'$: only those corresponding to deleted edges need be.

Theorem 1 has a number of implications. First, the necessity of Condition (4) will be exploited in the following section to provide an iterative method that searches for parameters that are a stationary point for the KL–divergence. Second, the sufficiency of Condition (4) implies that any method that searches for edge parameters, regardless of the criteria chosen, will yield parameters that are a stationary point for the KL–divergence, if the parameters give rise to exact marginals for variables corresponding to deleted edges. For example, if we search for parameters using ED-BP (Choi & Darwiche, 2006), and the parameters found lead to exact marginals, then these parameters will indeed be a stationary point for the KL–divergence.

Before we show how to identify parameters satisfying Condition (4), we note that parameters satisfying Condition (2) do not necessarily satisfy Condition (4) and, hence, are not necessarily a stationary point for $KL(Pr(.|\mathbf{e}), Pr'(.|\mathbf{e}'))$. We provide a simple network with four nodes in Appendix B demonstrating this point. Recall that Condition (2) characterizes IBP and some of its generalizations (Choi & Darwiche, 2006).

## 3.3 SEARCHING FOR PARAMETERS

Having characterized stationary points of the KL–divergence, we now proceed to develop an iterative procedure for finding a stationary point. Our method is based on the following result.

**Theorem 2** *Let $N$ be a Bayesian network and $N'$ be the network that results from deleting equivalence edges $U \to U'$. The edge parameters of $N'$ are a stationary point of $KL(Pr(.|\mathbf{e}), Pr'(.|\mathbf{e}'))$ if and only if:*

$$PM(u') = Pr(u|\mathbf{e}) \left( Pr'(\mathbf{e}') / \frac{\partial Pr'(\mathbf{e}')}{\partial \theta_{u'}} \right), \quad (5)$$

$$SE(u) = Pr(u|\mathbf{e}) \left( Pr'(\mathbf{e}') / \frac{\partial Pr'(\mathbf{e}')}{\partial \theta_{s'|u}} \right). \quad (6)$$

We have a number of observations about this theorem. First, if we have access to the true marginals $Pr(u|\mathbf{e})$, then this theorem suggests an iterative method that starts with some arbitrary values of parameters $PM(u')$ and $SE(u)$, leading to some initial approximate network $N'$. Using this network, we can compute the quantities $Pr'(\mathbf{e}')$, $\frac{\partial Pr'(\mathbf{e}')}{\partial \theta_{u'}}$ and $\frac{\partial Pr'(\mathbf{e}')}{\partial \theta_{s'|u}}$, which can then be used to compute new values for the parameters $PM(u')$ and $SE(u)$, one set at a time. The process can then be repeated until convergence (if at all). We will refer to this method as ED-KL, to be contrasted with ED-BP given earlier (Choi & Darwiche, 2006). Note that since the KL–divergence is non-negative, there exists a set of edge parameters that are globally minimal. However, a stationary point of the KL–divergence is not necessarily a global minima.

Second, the availability of the true marginals $Pr(u|\mathbf{e})$ typically implies that the network treewidth is small enough to permit the computation of these marginals. Hence, ED-KL is in general applicable to situations where the treewidth is manageable, but where the constrained treewidth is not. In these situations, the goal of deleting edges is to reduce the network constrained treewidth, making it amenable to algorithms that are exponential in constrained treewidth, such as MAP (Park & Darwiche, 2004), inference in credal networks (Cozman et al., 2004), and the computation of nonmyopic value of information (Krause & Guestrin, 2005).

Third, we have the following result which is critical for the practical application of ED-KL:

**Theorem 3** *Let $N$ be a Bayesian network and $N'$ be the network that results from deleting a single equivalence edge $U \to U'$. We then have*

$$Pr'(\mathbf{e}') = \sum_{uu'} \theta_{s'|u} \theta_{u'} \frac{\partial Pr(\mathbf{e})}{\partial \theta_{u'|u}},$$

*which implies:*

$$\frac{\partial Pr'(\mathbf{e}')}{\partial \theta_{u'}} = \sum_{u} \theta_{s'|u} \frac{\partial Pr(\mathbf{e})}{\partial \theta_{u'|u}},$$

$$\frac{\partial Pr'(\mathbf{e}')}{\partial \theta_{s'|u}} = \sum_{u'} \theta_{u'} \frac{\partial Pr(\mathbf{e})}{\partial \theta_{u'|u}}.$$

The main observation here is that $\partial Pr(\mathbf{e}) / \partial \theta_{u'|u}$ is a function of the original network and, therefore, is independent of the parameters $\theta_{u'}$ and $\theta_{s'|u}$—this is why the second and third equations above follow immediately from the first. Given the above equations, we can apply ED-KL to a single deleted edge, without the need for inference. That is, assuming that we have computed $\partial Pr(\mathbf{e}) / \partial \theta_{u'|u}$, we can use the above equations to compute updated values for edge parameters from the old values in constant time. This result will have implications in the following section, as we present a heuristic for deciding which edges to delete.

# 4 CHOOSING EDGES TO DELETE

Our method for deciding which edges to delete is based on scoring each network edge in isolation, leading to a total order on network edges, and then deleting edges according to the resulting order. For example, if we want to delete $k$ edges, we simply delete the first $k$ edges in the order.

The score for edge $U \to U'$ is based on the KL–divergence between the original network $N$ and an approximate network $N'$ which results from deleting the single edge $U \to U'$. The KL–divergence is computed using Lemma 1. This lemma requires some quantities from the original network $N$, which can be computed since the network is assumed to have a manageable treewidth. The lemma also requires that we have the parameters $\theta_{u'}$ and $\theta_{s'|u}$ for the deleted edge $U \to U'$, and the corresponding probability $Pr'(\mathbf{e}')$. These can be computed relatively easily using Theorem 3, assuming that we have computed $\partial Pr(\mathbf{e}) / \partial \theta_{u'|u}$ as explained in the previous section. Given these observations, all edge parameters, together with the corresponding KL scores, can be computed simultaneously for all edges using a single evaluation of the original network. Moreover, the computed parameters have another use beyond scoring edges: when used as initial values for ED-KL, they tend to lead to better convergence rates. We indeed employ this observation in our experiments.

# 5 EMPIRICAL ANALYSIS

We present experimental results in this section on a number of Bayesian networks, to illustrate a number of points on the relative performance of ED-KL and ED-BP. We start with Figure 4 which depicts the quality of computed approximations according to the *exact* KL–divergence[5]; see Equation 3. For each approximation scheme, we consider two methods for deleting edges. For ED-KL, we delete edges randomly (ED-KL-

[5]To compute the exact KL–divergence, see, for example, (Choi et al., 2005).

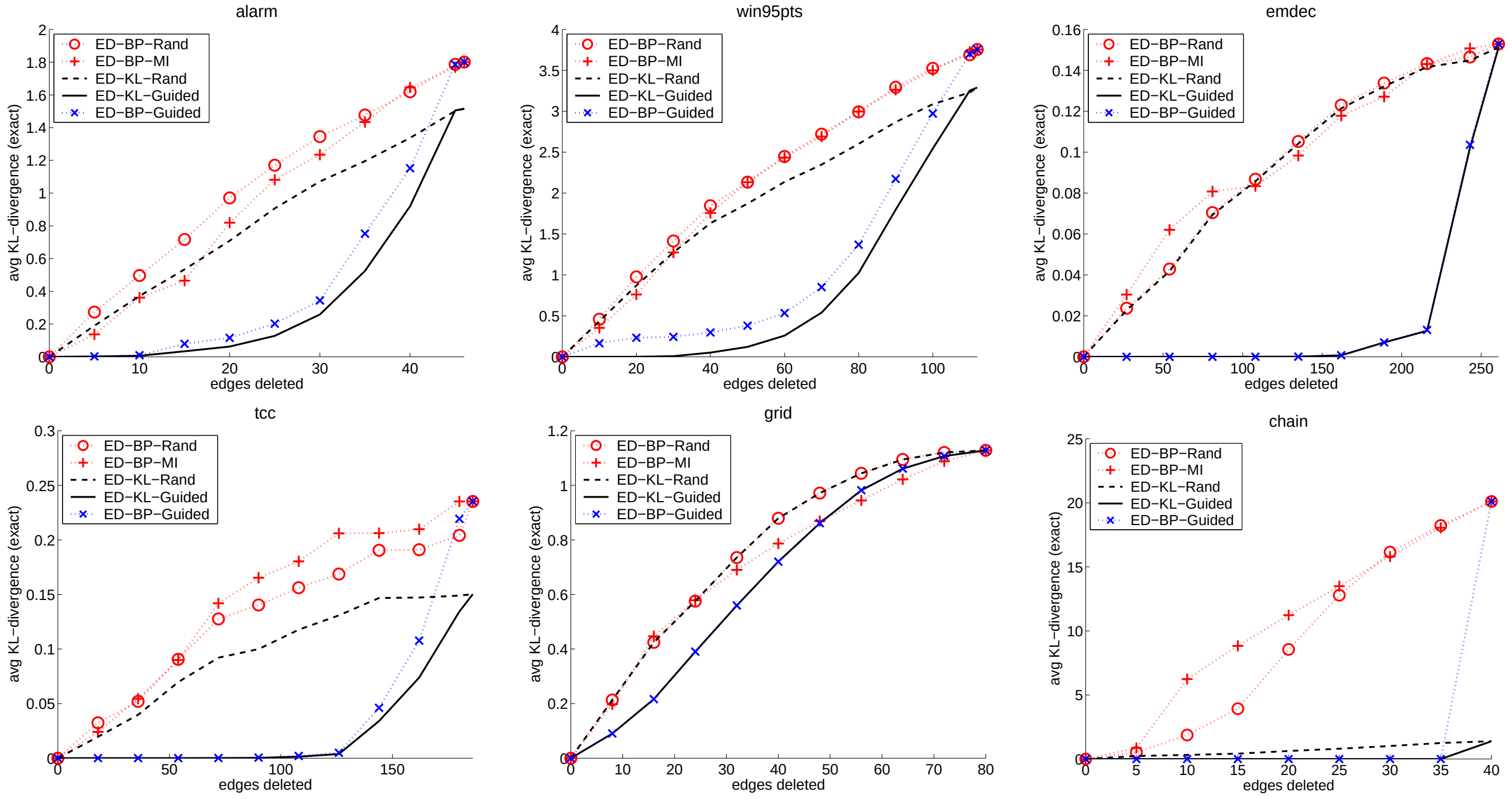

Figure 4: Comparing ED-KL and ED-BP using the <u>exact KL</u>, with two methods for edge deletion.

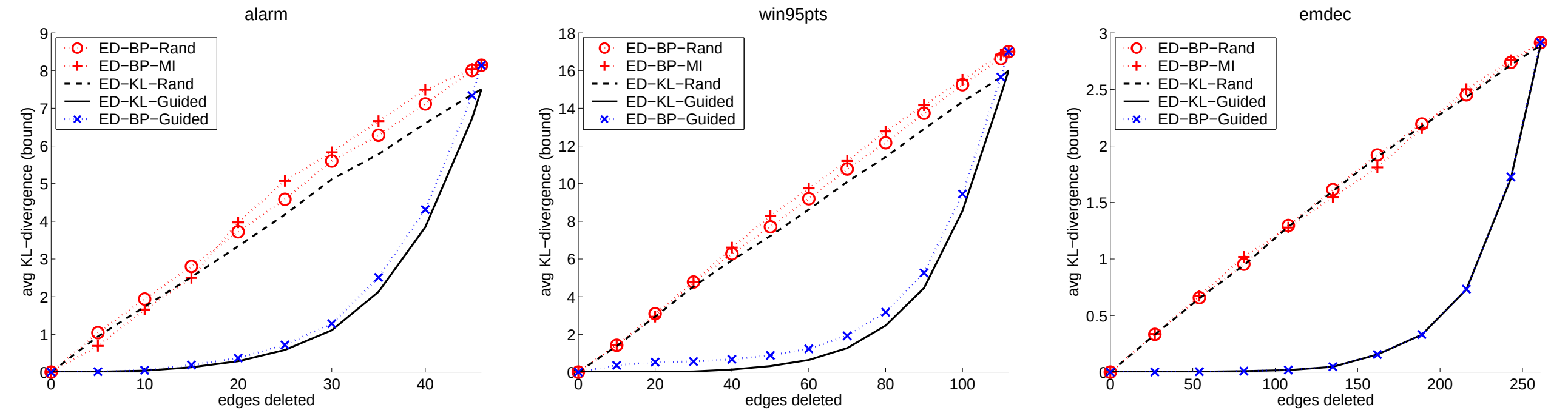

Figure 5: Comparing ED-KL and ED-BP using the <u>KL bound</u>, with two methods for edge deletion.

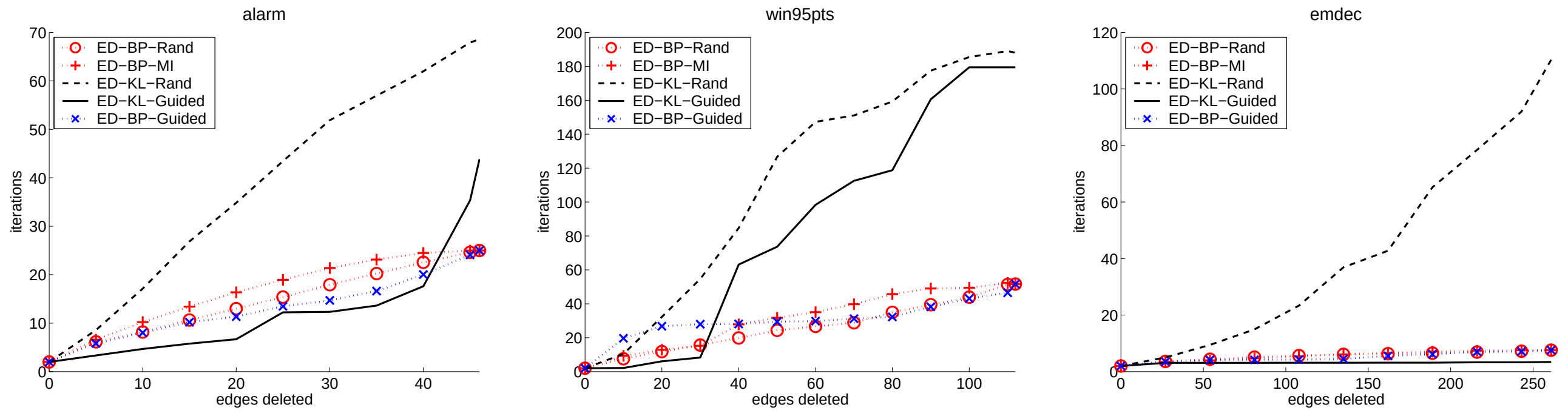

Figure 6: Comparing ED-KL and ED-BP according to <u>number of iterations</u> to converge.

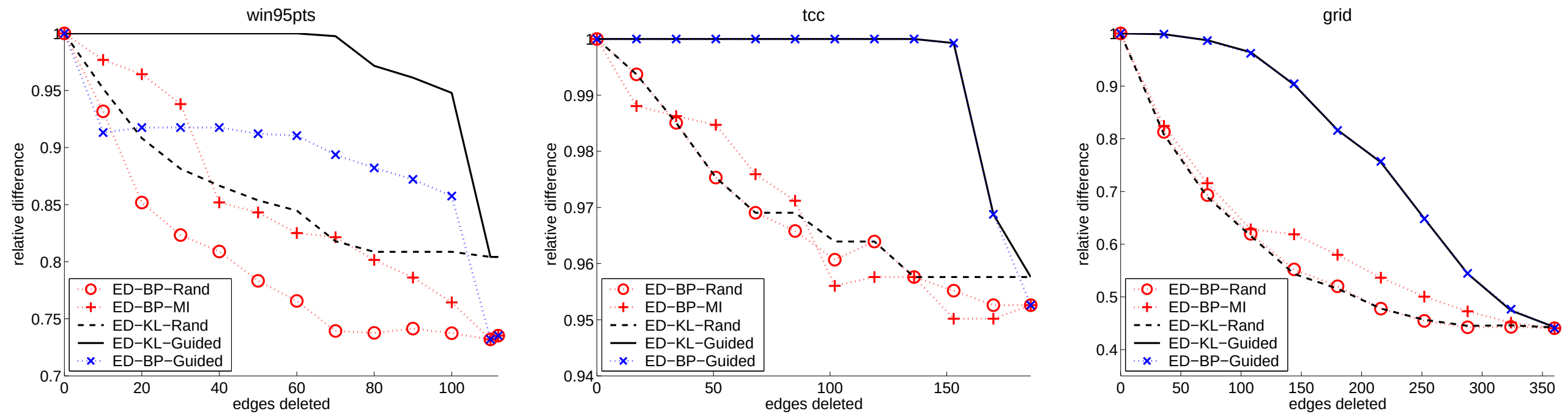

Figure 7: Approximating MAP using ED-KL and ED-BP.

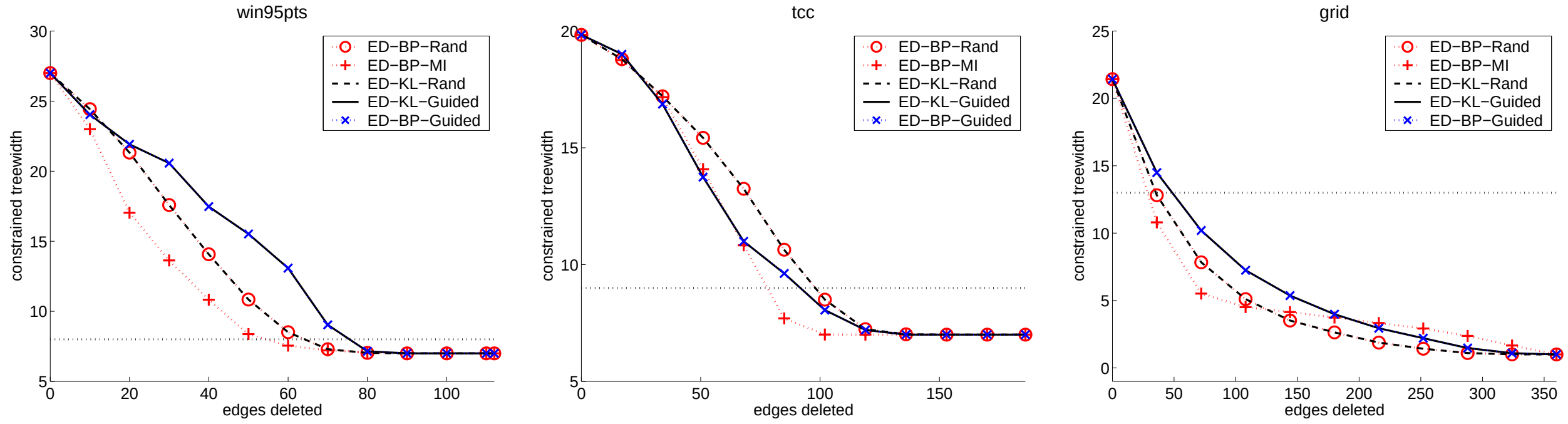

Figure 8: Reducing constrained treewidth by deleting edges. The horizontal line denotes network treewidth, as approximated using a min-fill heuristic.

RAND) and according to the heuristic of Section 4 (ED-KL-GUIDED). For ED-BP, we delete edges randomly (ED-BP-RAND) and according to a heuristic based on mutual information (ED-BP-MI) given in (Choi & Darwiche, 2006). As is clear from the figures, ED-KL is overwhelmingly superior as far as minimizing the KL–divergence, sometimes even using random deletion of edges.

Figure 5 depicts sample results for the quality of approximations according to the KL–divergence *bound;* see Equation 3. As mentioned earlier, ED-KL searches for stationary points of this bound instead of stationary points of the exact KL–divergence, yet empirically one does not see much difference between the two on these networks.

Figures 4 and 5 depict another approximation scheme, ED-BP-GUIDED, which deletes edges based on the heuristic of Section 4, but then uses ED-BP to search for parameters instead of ED-KL. This is a hypothetical method since the mentioned heuristic assumes that the network treewidth is manageable, a situation under which one would want to apply ED-KL instead of ED-BP. Yet, our results show that ED-BP is consistently very close to ED-KL in this case as far as minimizing the KL–divergence. This observation is critical as it highlights the great importance of heuristics for deleting edges. In particular, these results show that ED-BP can do quite well in terms of minimizing the KL–divergence if the right edges are deleted!

Figure 6 depicts sample results on the speed of convergence for both ED-KL and ED-BP, again using the different methods for edge deletion. In two of these networks, ED-KL consistently converges faster than ED-BP. In the three omitted figures, due to space limitations, ED-KL is also superior to ED-BP.

We consider also sample results from using approximations identified by ED-KL to approximate MAP. Figure 7 depicts the relative difference $p/q$, where $p$ is the value of the MAP solution found in the approximate network $N'$ and $q$ is the value of a MAP solution in the original network $N$. It is clear from the figure that ED-KL-GUIDED produces the superior approximations, and can provide accurate solutions even when many edges are deleted. Again, based on the hypothetical method ED-BP-GUIDED, we see that it is possible for ED-BP to produce good MAP approximations as well if the right edges are deleted.

Figure 8 highlights how effective deleting edges is in reducing the constrained treewidth (approximated using a min-fill heuristic), and thus how effective deleting edges is in reducing the complexity of computing MAP. We see that good approximations can be maintained even when the constrained treewidth is reduced to the network treewidth. When we further delete every network edge, we have a fully factorized approximation of MAP, where the constrained (and network) treewidth corresponds to the size of the largest network CPT.

The plots given in this section correspond to averages of at least 50 instances per data point, where each instance correspond to evidence over all leaf nodes drawn from the network joint. We have also experimented with evidence drawn randomly (not from the joint), leading to similar results. Networks *tcc* and *emdec* are courtesy of HRL Labs, LLC. The *grid* and *chain* networks are synthetic and available from the authors. Networks *alarm* and *win95pts* are available at *http://www.cs.huji.ac.il/labs/compbio/Repository.*

# 6 RELATED WORK

Many variational methods pose the problem of approximate inference as exact inference in some approximate model, often seeking to minimize the KL–divergence, but weighing it by the approximate distribution (e.g., Jordan, Ghahramani, Jaakkola, & Saul, 1999; Jaakkola, 2000; Wiegerinck, 2000; Geiger & Meek, 2005). One example is the mean–field method, where we seek to approximate a network $N$ by a fully disconnected $N'$ (Haft, Hofmann, & Tresp, 1999). If we delete all edges from the network and try to parametrize edges using ED-KL, we would be solving the same problem solved by mean–field, except that our KL–divergence is weighted by the true distribution, leading to more expensive update equations. Other variational approaches typically assume particular structures in their approximate models, such as chains (Ghahramani & Jordan, 1997), trees (Frey, Patrascu, Jaakkola, & Moran, 2000; Minka & Qi, 2003), or disconnected subnetworks (Saul & Jordan, 1995; Xing, Jordan, & Russell, 2003). In contrast, ED-KL works for any network structure which is a subset of the original. In fact, the efficient edge deletion heuristic of Section 4 tries to select the most promising subnetworks and is quite effective as illustrated earlier. Again, most of these approaches weigh the KL–divergence by the approximate distribution for computational reasons, with the notable exceptions of (Frey et al., 2000; Minka & Qi, 2003).

Other methods of edge deletion have been proposed for Bayesian networks (Suermondt, 1992; Kjærulff, 1994; van Engelen, 1997), some of which can be rephrased using a variational perspective. All of these approaches, however, approximate a network independent of the given evidence, which is a dramatic departure from ED-KL and can lead to much worse behavior for less likely evidence. That is, these approaches approximate a network once for all queries, while ED-KL can approximate a network for each specific query.

# 7 CONCLUSION

We proposed a method, ED-KL, for approximating Bayesian networks by deleting edges from the original network and then finding stationary points for the KL–divergence between the original and approximate networks (while weighing the divergence by the true distribution). We also proposed an efficient heuristic for deciding which edges to delete from a network, with the aim of choosing network substructures that lead to high quality approximations.

The update equations of ED-KL require exact posteriors from the original network. This means that ED-KL is, in general, applicable to problems that remain hard even when treewidth is manageable, including MAP, nonmyopic value of information, and inference in credal networks. This is to be contrasted with our earlier method ED-BP, which updates parameters differently, coinciding with IBP and some of its generalizations.

Our empirical results provide good evidence to the quality of approximations returned by ED-KL, especially when compared to the approximations returned by ED-BP. Moreover, our results, both theoretical and empirical, shed new and interesting light on ED-BP (and, hence, IBP and some of its generalizations), showing that it can also produce high quality approximations (from a KL–divergence viewpoint), when deleting the right set of network edges.

**Acknowledgments**

This work has been partially supported by Air Force grant #FA9550-05-1-0075-P00002 and by JPL/NASA grant #1272258.

# A Proof Sketches

Note that $u \sim w$ signifies that $u$ and $w$ are compatible instantiations.

**Proof of Lemma 1** Deleting edges $U \to U'$, we have:

$$KL(Pr(.|\mathbf{e}), Pr'(.|\mathbf{e}')) = \sum_w Pr(w|\mathbf{e}) \log \frac{Pr(w|\mathbf{e})}{Pr'(w|\mathbf{e}')}$$

$$= \sum_w Pr(w|\mathbf{e}) \log \frac{Pr(w,\mathbf{e})}{Pr'(w,\mathbf{e}')} + \log \frac{Pr'(\mathbf{e}')}{Pr(\mathbf{e})}$$

$$= \sum_w Pr(w|\mathbf{e}) \log \prod_{uu' \sim w} \frac{\theta_{u'|u}}{\theta_{u'}\theta_{s'|u}} + \log \frac{Pr'(\mathbf{e}')}{Pr(\mathbf{e})}$$

$$= \sum_{w} \sum_{uu' \sim w} Pr(w|\mathbf{e}) \log \frac{\theta_{u'|u}}{\theta_{u'}\theta_{s'|u}} + \log \frac{Pr'(\mathbf{e}')}{Pr(\mathbf{e})}$$

$$= \sum_{U \to U'} \sum_{uu'} \sum_{w \models uu'} Pr(w|\mathbf{e}) \log \frac{\theta_{u'|u}}{\theta_{u'}\theta_{s'|u}} + \log \frac{Pr'(\mathbf{e}')}{Pr(\mathbf{e})}$$

$$= \sum_{U \to U'} \sum_{uu'} Pr(uu'|\mathbf{e}) \log \frac{\theta_{u'|u}}{\theta_{u'}\theta_{s'|u}} + \log \frac{Pr'(\mathbf{e}')}{Pr(\mathbf{e})}$$

$$= \sum_{U \to U'} \sum_{u=u'} Pr(uu'|\mathbf{e}) \log \frac{1}{\theta_{u'}\theta_{s'|u}} + \log \frac{Pr'(\mathbf{e}')}{Pr(\mathbf{e})}.$$

The last equation follows, since when $u$ does not agree with $u'$, we have that $Pr(uu'|\mathbf{e}) \log \theta_{u'|u} = 0 \log 0$, which we assume is equal to zero, by convention. □

**Proof of Theorem 1** Note that when $u = u'$, we have $Pr(uu'|\mathbf{e}) = Pr(u|\mathbf{e}) = Pr(u'|\mathbf{e})$.

*First direction of theorem.* Let $f$ be the KL–divergence as given in Lemma 1. Setting $\partial f / \partial \theta_{u'}$ to zero, we get:

$$\frac{\partial f}{\partial \theta_{u'}} = -\frac{Pr(u|\mathbf{e})}{\theta_{u'}} + \frac{1}{Pr'(\mathbf{e}')} \frac{\partial Pr'(\mathbf{e}')}{\partial \theta_{u'}} = 0, \quad (7)$$

where $u$ agrees with $u'$. We then have

$$Pr(u|\mathbf{e}) = \frac{\theta_{u'}}{Pr'(\mathbf{e}')} \frac{\partial Pr'(\mathbf{e}')}{\partial \theta_{u'}} = \frac{Pr'(u', \mathbf{e}')}{Pr'(\mathbf{e}')} = Pr'(u'|\mathbf{e}')$$

Similarly, to show $Pr(u|\mathbf{e}) = Pr'(u|\mathbf{e}')$. Note that constraints such as normalization are inactive here.

*Second direction of theorem.* Given a network $N'$ where marginals on $U$ and $U'$ are exact, we want to verify that the edge parameters are stationary points. If we take the partial derivative with respect to $\theta_{u'}$:

$$\begin{aligned} \frac{\partial f}{\partial \theta_{u'}} &= \frac{1}{\theta_{u'}} \left( -Pr(u|\mathbf{e}) + \frac{\theta_{u'}}{Pr'(\mathbf{e}')} \frac{\partial Pr'(\mathbf{e}')}{\partial \theta_{u'}} \right) \\ &= \frac{1}{\theta_{u'}} \left( -Pr(u|\mathbf{e}) + \frac{Pr'(u', \mathbf{e}')}{Pr'(\mathbf{e}')} \right) \\ &= \frac{1}{\theta_{u'}} \left( -Pr(u|\mathbf{e}) + Pr'(u'|\mathbf{e}') \right). \end{aligned}$$

We are given $Pr'(u'|\mathbf{e}') = Pr(u|\mathbf{e})$, thus $\partial f / \partial \theta_{u'} = 0$. Similarly, to show $\partial f / \partial \theta_{s'|u} = 0$. □

**Proof of Theorem 2** Equation 5 follows easily from Equation 7. Equation 6 follows from $\partial f / \partial \theta_{s'|u}$. □

**Proof of Theorem 3** First, we have:

$$Pr'(\mathbf{e}') = \sum_{uu'} Pr'(uu', \mathbf{e}')$$

$$= \sum_{uu'} \theta_{s'|u}\theta_{u'} \frac{\partial^2 Pr'(\mathbf{e}')}{\partial \theta_{s'|u} \partial \theta_{u'}} = \sum_{uu'} \theta_{s'|u}\theta_{u'} \frac{\partial Pr'(u, \mathbf{e})}{\partial \theta_{u'}}.$$

Note that the distribution induced by a network where a single edge $U \to U'$ has been deleted is equivalent to the distribution induced by another network $N'$ that is identical in structure to $N$, except that $\theta_{u'|u} = \theta_{u'}$ for all $u$. We then have:

$$Pr'(\mathbf{e}') = \sum_{uu'} \theta_{s'|u}\theta_{u'} \frac{\partial Pr'(\mathbf{e}')}{\partial \theta_{u'|u}} = \sum_{uu'} \theta_{s'|u}\theta_{u'} \frac{\partial Pr(\mathbf{e})}{\partial \theta_{u'|u}}.$$

The other relations follow easily. □

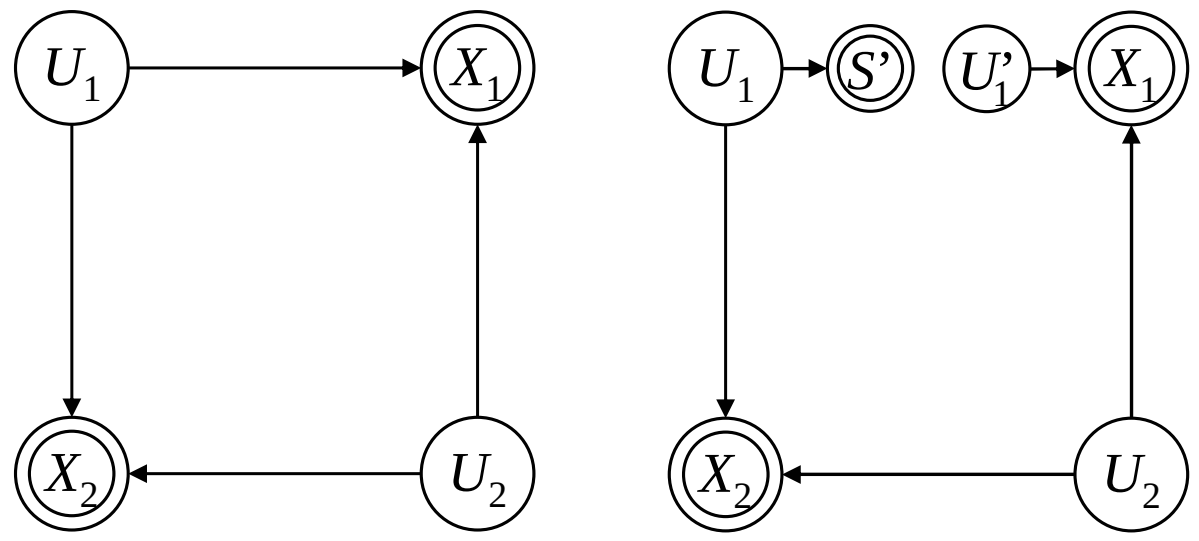

Figure 9: An example network $N$ (left) where deleting a single edge (right) may have infinitely many ED-BP fixed points.

## B Example

We demonstrate here an example where ED-BP fixed points are not necessarily stationary points of the KL–divergence. This example also shows that even if we delete a single edge, ED-BP can have infinitely many parametrizations satisfying Condition (2), even though there exists an ED-BP (and ED-KL) fixed point minimizing the *KL bound*, $KL(Pr(.|\mathbf{e}), Pr'(.|\mathbf{e}'))$ (as well as minimizing the *exact KL*). This example also corresponds to an instance of IBP (with a particular message passing schedule), since edge deletion renders the network a polytree (Choi & Darwiche, 2006).

Our example is depicted in Figure 9. Variables $U_i$ have parameters $\theta_{u_i} = \theta_{\bar{u}_i} = 0.5$. Variables $X_j$ are fixed to states $x_j$, and assert the equivalence of $U_1$ and $U_2$: $\theta_{x_j|u_1 u_2} = 1$ iff $u_1 = u_2$. Conditioning on evidence $\mathbf{e} = x_1 x_2$, we have $Pr(u_1|\mathbf{e}) = Pr(u_2|\mathbf{e}) = 0.5$. If we delete the edge $U_1 \to X_1$ (implicitly, we delete an edge $U_1 \to U_1'$), then any non-zero parameterization of our edge parameters satisfies the ED-BP fixed point conditions given by Condition (2). For example, when $\theta_{s'|u_1} = \theta_{s'|\bar{u}_1} = 0.5$, and $\theta_{u_1'} = \theta_{\bar{u}_1'} = 0.5$, the KL–divergence is zero and thus minimized, yielding parent and clone marginals that are exact. By Theorem 1, edges parameters are then a stationary point of the KL–divergence. However, when $\theta_{u_1'} \neq 0.5$, the parent and clone marginals are not exact, and thus edges parameters are not a stationary point of the KL–divergence, again by Theorem 1.